# MEC: Memory-efficient Convolution for Deep Neural Network

Minsik Cho [1]  Daniel Brand [1]

## Abstract

Convolution is a critical component in modern deep neural networks, thus several algorithms for convolution have been developed. Direct convolution is simple but suffers from poor performance. As an alternative, multiple indirect methods have been proposed including im2col-based convolution, FFT-based convolution, or Winograd-based algorithm. However, all these indirect methods have high memory-overhead, which creates performance degradation and offers a poor trade-off between performance and memory consumption. In this work, we propose a memory-efficient convolution or MEC with compact lowering, which reduces memory-overhead substantially and accelerates convolution process. MEC lowers the input matrix in a simple yet efficient/compact way (i.e., much less memory-overhead), and then executes multiple small matrix multiplications in parallel to get convolution completed. Additionally, the reduced memory footprint improves memory sub-system efficiency, improving performance. Our experimental results show that MEC reduces memory consumption significantly with good speedup on both mobile and server platforms, compared with other indirect convolution algorithms.

## 1. Introduction

Deep neural network (DNN) consists of many layers to perform a task such as image classification/recognition, speech recognition, natural language translation, and so on. Among these layers, the convolution layer is one of the most important, but the slowest and most memory-intensive ones in advanced/modern convolutional DNN (Abuzaid et al., 2015; Chen et al., 2016; Cong & Xiao, 2014; Denton et al., 2014; Park et al., 2016a; Vasilache et al., 2014). To address the performance issues in convolutional layers, efficient/approximation algorithms have been proposed (Chellapilla et al., 2006; Denton et al., 2014; Jaderberg et al., 2014; Jia, 2014; Vasilache et al., 2014), tailed implementations for limited cases have been actively investigated (Lavin, 2015), and industrial-strength libraries are offered (Chetlur et al., 2014).

However, the previous approaches have not directly addressed the memory consumption problem. This is becoming a critical issue as DNNs are getting in end-point devices with limited memory (e.g., mobile/IOT devices) (Chen et al., 2015; Collins & Kohli, 2014; Gong et al., 2014; Kim et al., 2015; Lebedev et al., 2014; Wang & Cheng, 2016) so as to minimize response delay (e.g., better user experience) and network overhead (Han et al., 2015; Lane et al., 2016; 2015). On the other hand, the reduced memory consumption leads to smaller SRAM usage, which can save energy consumption (e.g., leakage current) on mobile devices (Park et al., 2015). Moreover, memory footprint itself has critical impact on convolution computation efficiency (Li et al., 2016; Park et al., 2016b). Therefore, minimizing memory footprint in convolution is critical for future deep-learning applications on wide variety of devices and platforms.

In this paper, we propose a new memory-efficient convolution algorithm, MEC which can reduce memory-overhead and further improve the performance of computing convolution in DNN. MEC uses a simple yet novel way of lowering the input matrix in a highly compact way, while still exploiting fast matrix-matrix multiplication available in a highly-optimized package such as BLAS (Jia, 2014). The reduced memory footprint improves memory sub-system efficiency (i.e., improves cache locality), so that MEC accelerates the convolution computation itself without compromising accuracy. Through extensive experiments on both mobile and server platforms with CPU/GPU, we show that MEC can be a very generic/efficient algorithm suitable to various platforms with memory constraints. Further, the key ideas in MEC should be beneficial/complementary to any variant of conventional im2col-based convolution by reducing either memory consumption or memory-bus traffic (i.e., less traffic from global memory to shared memory on GPU) (Chellapilla et al., 2006; Chetlur et al., 2014; Jia,

[1]IBM T. J. Watson Research Center, NY, USA. Correspondence to: Minsik Cho <minsikcho@us.ibm.com>.

*Table 1.* Notations.

| | | |
|---|---|---|
| $a:b$ | SEQUENCE | $\{a, a+1, ... b-1\}$ |
| $A[a,b]$ | MATRIX ELEMENT | |
| $A[a:b,c:d]$ | SUB-MATRIX | $A[i,j], i \in a:b, j \in c:d$ |
| $I$ | INPUT TENSOR | $i_n \times i_h \times i_w \times i_c$ |
| $K$ | KERNEL TENSOR | $k_h \times k_w \times i_c \times k_c$ |
| $O$ | OUTPUT TENSOR | $i_n \times o_h \times o_w \times k_c$ |
| $L$ | LOWERED TENSOR | $i_n \times o_w \times i_h \times k_w \times i_c$ |
| $s_h, s_w$ | KERNEL STRIDE | |

2014).

The rest of the paper is organized as follows. We review related works and present preliminaries in Section 2. Section 3 presents our proposed algorithm, MEC. Experimental results are in Section 4. Section 5 concludes this paper.

## 2. Preliminaries

### 2.1. Notations

Notation used in this paper is listed in Table 1. For integers we use small letters, for tensors and matrices we use capital letters. We adopt the C-language convention as representing tensors and matrices in row-major order. For example, a $p \times q \times r$ tensor is an array of $pqr$ elements. The array can be interpreted as consisting of $p$ sections, each divided into $q$ subsections, each having $r$ elements. The same array can also be interpreted as $p \times qr$ matrix, or as $pq \times r$ matrix, etc. We specifically interpret a tensor as a matrix when it requires matrix operations, otherwise (i.e., for data movement) we keep the tensor form. If we work with a math library, such as cuBLAS (cuBLAS), which requires column-major order, then we still use the same row-major representation, but interpret all matrices as being transposed.

We use the notation $a:b$ to denote a sub-matrix. Thus, an $m \times n$ matrix could be written as $A[0:m, 0:n]$. The most common form of a sub-matrix will be of the form $A[i:i+p, j:j+q]$. It is a $p \times q$ sub-matrix with top left corner at the element $A[i,j]$, which can be easily represented in the BLAS interface without moving any elements by having leading dimension $ld = n$.

The subject of this paper is 2-dimensional convolution $O = I \star K$ with strides $s_h, s_w$. For simplicity of explanation any padding with zeroes is assumed to have been already applied to the input $I$. The output matrix $O$ will have the dimensions

$$o_{h,w} = \frac{i_{h,w} - k_{h,w}}{s_{h,w}} + 1 \qquad (1)$$

### 2.2. Previous Work

Due to the importance of DNN, several techniques for efficient convolution computation have been proposed (Chetlur et al., 2014; Perkins, 2016). The most relevant to our work is im2col-based convolution, FFT (Fast Fourier Transform)-based convolution (Highlander & Rodriguez, 2016; Mathieu et al., 2013; Vasilache et al., 2014), and Winograd-based convolution (Lavin, 2015). MEC provides the same functionality with reduced memory requirements.

- im2col-based convolution transforms/lowers the input matrix into a Toeplitz matrix with redundancy (a.k.a, lowered matrix) such that convolution can be performed as fast matrix-matrix multiplication, which can take advantage of highly optimized linear algebra packages including BLAS (Chellapilla et al., 2006; Chetlur et al., 2014; Jia, 2014).

- FFT-based convolution relies on the fact that convolution can be done as simple multiplication in the frequency domain. However, FFT-based convolution incurs memory-overhead because all the kernels must be padded to be at the same size as the input matrix. Thus, memory-overhead becomes really high when kernels are relatively smaller (e.g., 3x3) than input matrices (Chetlur et al., 2014; He et al., 2015; Perkins, 2016; Simonyan & Zisserman, 2014).

- Winograd-based convolution is based on the Coppersmith-Winograd algorithm (Winograd, 1980) which shows how to reduce multiplication counts at a cost of more addition counts and a large number of intermediate products. It is shown in (Lavin, 2015; Park et al., 2016a) that Winograd-based convolution can be efficient for small kernels on GPU.

In contrast to the above schemes, which do not degrade accuracy, various approximation strategies have been proposed including low-rank/monochromatic approximation (Denton et al., 2014; Jaderberg et al., 2014), vector quantization (Gong et al., 2014), fine-tuning (Lebedev et al., 2014), and DCT (Discrete Cosine Transform)/hashing (Lebedev et al., 2014).

## 3. Algorithm

In this section, we propose our algorithm for convolution, MEC, with detailed examples. The main goal of MEC is to reduce memory-overhead during convolution, which can be beneficial for any convolutional DNN in three aspects:

- MEC can enable training or inferencing with a larger model for a given memory capacity.

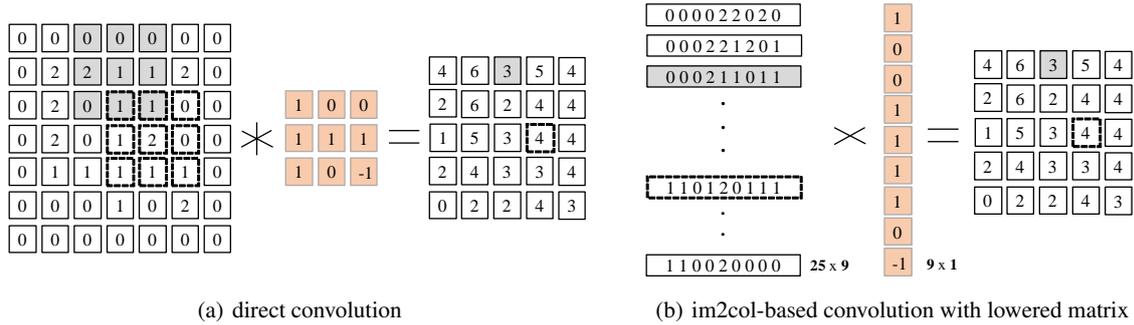

Figure 1. Conventional convolution examples with $i_w = i_h = 7, k_h = k_w = 3, s_h = s_w = 1, o_w = o_h = 5$ ($i_n = i_c = k_c = 1$).

- MEC can allow larger mini-batch sizes to speedup turn-around/per-epoch-latency during training.
- MEC can accelerate computation by improving memory sub-system efficiency (e.g. more cache hits).

In contrast to the widely-adopted im2col-based convolution (Chellapilla et al., 2006; Chetlur et al., 2014; Jia, 2014), MEC performs compact/BLAS-friendly lowering such that memory-overhead can be minimized without degrading performance/accuracy. Section 3.1 motivates MEC, and Section 3.2 highlights the key idea in MEC. Section 3.3 formally presents MEC with implementation details.

### 3.1. Motivation

In this section, we review im2col-based convolution and its pros and cons with Fig. 1 which sketches direct convolution in (a) and im2col-based convolution using BLAS in (b). In direct convolution, one element of the output matrix $O$ is produced by a dot-product between the kernel $K$ and a sub-matrix of the input $I$. The sub-matrices are obtained by sliding $K$ over $I$ in both dimensions. Each subsequent sub-matrix is obtained by sliding the distance $s_h$ or $s_w$, respectively. For example, Fig. 1 (a) shows two sub-matrices in gray and dotted boxes w.r.t. the $3 \times 3$ kernel are processed to generate the corresponding output values in gray and dotted boxes (i.e., 3 and 4), respectively.

Direct convolution is simple and straightforward without memory-overhead. However, it is known that the same convolution can be done more efficiently with a lowered matrix (a.k.a. *im2col*) and gemm in BLAS (Chellapilla et al., 2006; Chetlur et al., 2014; Jia, 2014) by off-loading the geometry-specific specializations in convolution to a plain matrix, which is depicted in Fig. 1 (b). Specifically, each sub-matrix instance w.r.t. $K$ is linearized into a row of the lowered matrix $L$ as in (b). For example, the gray and dotted sub-matrices in (a) are transformed into the gray and dotted rows in (b), respectively. Then the output matrix $O = L \times K$, can be computed efficiently by optimized libraries (cuBLAS; Kågström et al., 1998; MKL; OpenBLAS). im2col-based convolution is generic enough to be used in any DNN on both mobile/IoT and high-end platforms (Chetlur et al., 2014; Lane et al., 2015).

The major drawback of im2col-based convolution is that it comes with memory-overhead of temporarily storing the lowered matrix $L$ with dimension

$$i_n o_h o_w \times k_h k_w k_c \qquad (2)$$

which shows that the memory requirement grows quadratically with problem size. The example in Fig. 1 (b) shows that the lowered matrix has size $25 \times 9$, which is even lager than the original input matrix. MEC mainly aims to perform the same convolution yet with less memory-overhead, while improving computational efficiency.

### 3.2. MEC Overview

In this section, we highlight the key idea in our memory-efficient convolution algorithm, MEC based on a compact lowering scheme. The main reason why the im2col-based algorithm has large memory-overhead is because there is a significant amount of redundancy in the lowered matrix when $s_h$ or $s_w$ is small and $K$ is large. And, the overhead becomes even worse when $K$ is relatively smaller than $I$ which occurs frequently in the state-of-the-art DNN architectures (He et al., 2015; Perkins, 2016; Simonyan & Zisserman, 2014; Szegedy et al., 2014). In order to reduce memory-overhead, therefore, it is critical to reduce the amount of redundancy in the lowered matrix and keep the computation pattern BLAS-compatible (otherwise, the poor computation itself may slow down the entire convolution).

MEC overcomes such challenges by lowering multiple columns at once rather than each single individual sub-matrix w.r.t. $K$. Consider the example in Fig. 2 for key ideas and details. MEC copies sub-matrices $W$ (shaded in Fig. 2) of size $i_h \times k_w$ (which is $7 \times 3$) into one row of $L$.

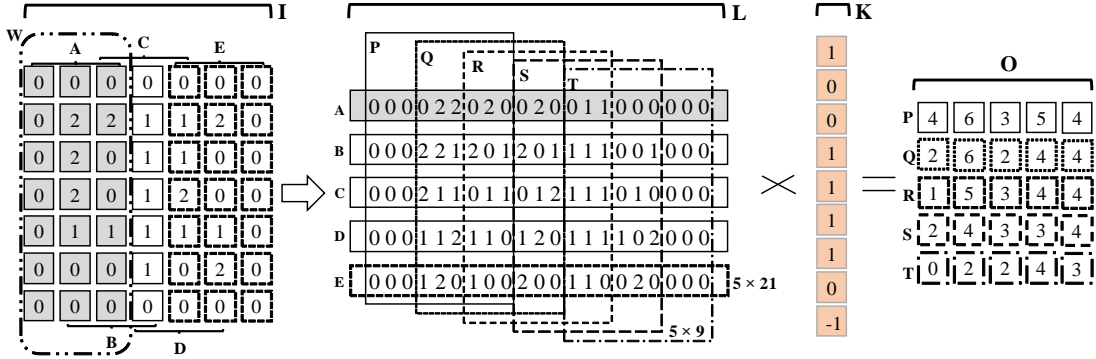

*Figure 2.* MEC example for the same problem in Fig. 1

For example, $A$ is the first partition of $I$, $A = I[0:7, 0:3]$. Then, we slide $W$ by $s_w$ (which is 1) to the right and create another partition $B = I[0:7, 1:4]$. As we continue this process in Fig. 2, there will be 5 *horizontal* partitions, $\{A, B, C, D, E\}$ in $L$ eventually. The resulting lowered matrix, $L$ has dimensions $5 \times 21$, which is 54% smaller than one in Fig. 1 with dimensions $25 \times 9$.

Once the lowered matrix $L$ is formed, MEC multiplies $L$ by $K$ in a way significantly different from im2col-based algorithms. MEC creates another set of *vertical* partitions, $\{P, Q, R, S, T\}$ within $L$, where each partition is of size of $o_w \times k_h k_w$ (which is $5 \times 9$). Each subsequent partition is obtained by shifting to the right by $s_h k_w$ (which is 3) elements. For example, $P = L[0:5, 0:9]$ and $Q = L[0:5, 3:12]$. Then each row of the output matrix $O$ is the product between one of the partitions in $\{P, Q, R, S, T\}$ and K. Rows in $O$ in Fig. 2 are annotated with the corresponding source partitions.

These multiplications rely on the BLAS *gemm* interface in three ways. First, the $k_h \times k_w$ matrix $K$ is interpreted as a $k_h k_w \times 1$ matrix. Second, the partitions $\{P, Q, R, S, T\}$ are specified by providing a pointer to the initial element and $ld = i_h k_w$, which is the entire length of one row of $L$. Thirdly, each row of $O$ is formed by 5 separate *gemm* calls between $\{P, Q, R, S, T\}$ and $K$. Although the number of *gemm* calls increases, the total number of mult/add operations remains identical to that of the im2col-based convolution, keeping computationally complexity same.

Intuitively, MEC eliminates the *vertical* redundancy in the conventional im2col-based convolution. Then it *recovers* the information by merely shifting the vertical partitions (i.e., $P, Q, R, S, T$) by a constant interval. These sub-matrix manipulations are made efficient by keeping the pattern BLAS compatible. The lowering in MEC is highly efficient as we move fewer elements from $I$ to *smaller* $L$,

**Algorithm 1** $O = VanillaMEC(I, K, s)$
1: Allocate $O$ with $o_h o_w$ elements
2: Allocate $L$ with $o_w i_h k_w$ elements
3: Interpret $L$ as $o_w \times i_h \times k_w$ tensor
4: **for** $w \in 0 : o_w, h \in 0 : i_h$ **in parallel do**
5:     $L[w, h, 0 : k_w] = I[h, s_w w : s_w w + k_w]$
6: **end for**
7: Interpret $L$ as $o_w \times i_h k_w$ matrix
8: Interpret $K$ as $k_h k_w \times 1$ matrix
9: Interpret $O$ as $o_h \times o_w$ matrix
10: **for** $h \in 0 : o_h$ **in parallel do**
11:     $O[h, 0 : o_w] =$
       $L[0 : o_w, s_h k_w h : s_h k_w h + k_h k_w] \times K$
12: **end for**
13: Return $O$

compared with im2col-based convolution, saving memory-bus traffic as well.

The process is stated in Algorithm 1 where $i_n = i_c = k_c = 1$. It first allocates the output $O$ and temporary $L$. The first loop in line 4 forms the matrix $L$, which copies $k_w$ consecutive elements from $I$ to $L$, and all these copies can be done in parallel. The second loop in line 10 forms the output $O$. Each execution of the body is done by one gemm call, and those matrix multiplications can be parallelized.

### 3.3. MEC Algorithm

In this section, we present the complete MEC by extending Algorithm 1 to Algorithm 2 in order to handle channels ($i_c$ and $k_c$) and mini-batches ($i_n$), and discuss the implementation details in the context of deep-learning (mainly about image format issue). Due to the compact lowering in MEC, it is computationally advantageous to use $I$ in $i_n \times i_h \times i_w \times i_c$ (or n-h-w-c) as in Table 2, because it ensures vertical redundant pixels to be eliminated and re-covered in a contiguous memory space.

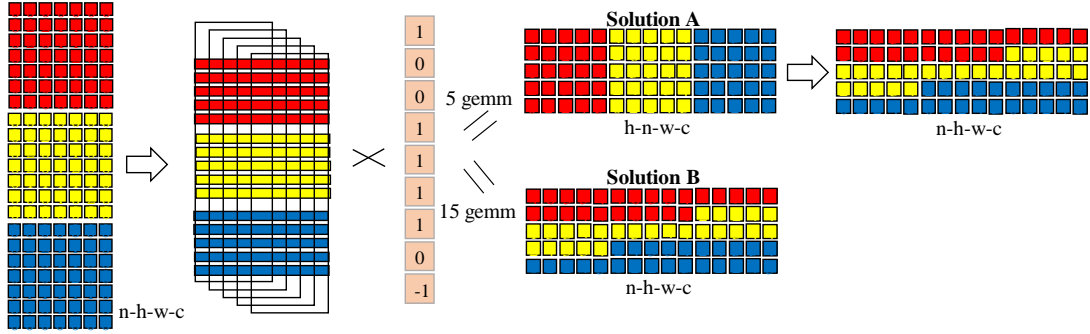

*Figure 3.* MEC with mini-batch example

**Algorithm 2** $O = MEC(I, K, s)$
1: Allocate $O$ with $i_n o_h o_w k_c$ elements
2: Allocate $L$ with $i_n o_w i_h k_w i_c$ elements
3: Interpret $L$ as $i_n \times o_w \times i_h \times k_w \times i_c$ tensor
4: **for** $n \in 0 : i_n, w \in 0 : o_w, h \in 0 : i_h$ **in parallel do**
5:     $L[n, \ w, \ h, \ 0 : k_w, \ 0 : i_c] =$
      $I[n, \ h, \ s_w w : s_w w + k_w, \ 0 : i_c]$
6: **end for**
7: Interpret $K$ as $k_h k_w i_c \times k_c$ matrix
8: **if** $o_w \leq T$ and $|O| \leq |L|$ **then**
9:     Interpret $L$ as $i_n o_w \times i_h k_w i_c$ matrix
10:    Interpret $O$ as $o_h \times i_n o_w k_c$ matrix
11:    **for** $h \in 0 : o_h$ **in parallel do**
12:       $O[h, \ 0 : i_n o_w k_c] =$
        $L[0 : i_n o_w, \ s_h k_w i_c h : s_h k_w i_c h + k_h k_w i_c] \times K$
13:    **end for**
14:    Copy $L = O$
15:    Interpret $L$ as $o_h \times i_n \times o_w k_c$ tensor
16:    Interpret $O$ as $i_n \times o_h \times o_w k_c$ tensor
17:    **for** $n \in 0 : i_n, h \in 0 : o_h$ **in parallel do**
18:       $O[n, h, 0 : o_w k_c] = L[h, n, 0 : o_w k_c]$
19:    **end for**
20: **else**
21:    Interpret $L$ as $i_n$ matrices of $o_w \times i_h k_w i_c$
22:    Interpret $O$ as $i_n$ matrices of $o_h \times o_w k_c$
23:    **for** $n \in 0 : i_n, h \in 0 : o_h$ **in parallel do**
24:       $O[n][h, \ 0 : o_w k_c] =$
        $L[n][0 : o_w, \ s_h k_w i_c h : s_h k_w i_c h + k_h k_w i_c] \times K$
25:    **end for**
26: **end if**
27: Return $O$ as $i_n \times o_h \times o_w k_c$ tensor

Based on $I$ as $i_n \times i_h \times i_w \times i_c$, Algorithm 2 still has the same key idea in presence of channels and mini-batches. The lowering step lines 4-6 in Algorithm 1 is similar to lines 4-6 in Algorithm 2. However, the parallel multiplication loop in lines 10-12 in Algorithm 1 extends to lines 8-25 in Algorithm 2 mainly due to the image format issue.

A direct extension of Algorithm 1 would interpret $O$ as $o_h \times i_n o_w k_c$ matrix, and perform $o_h$ multiplications for convolution of the whole mini-batch. This leads to the output format `h-n-w-c`, which is different from the input format of $I$. This may be acceptable in DNNs, where each convolution layer is followed by a pooling layer expecting `h-n-w-c` format and generating the standard `n-h-w-c` format. However, it would be troublesome in a network where all layers expect and produce the `n-h-w-c` format. Therefore, we provide two solutions depicted in Fig. 3 to handle such format-related issues.

**Solution A** (Lines 9 to 19 of Algorithm 2) First we perform the direct extension of Algorithm 1 (lines 9 - 13) and end up with $O$ in format `h-n-w-c`. Then, we transform $O$ into `n-h-w-c` format (lines 14-19) where we repurpose $L$ as an auxiliary space.

**Solution B** (lines 21 to 25 of Algorithm 2) We can handle the $i_n$ samples in the mini-batch separately as in line 21, resulting in $i_n o_h$ parallel/batched gemm calls with smaller inputs, as opposed to $o_h$ *gemm* calls with larger inputs. This will directly generate $O$ in `n-h-w-c`.

In terms of complexity, both solutions perform the same number of floating point multiplications. In practice, however, the size of sub-matrices can impact performance, particularly on implementation-sensitive platform like GPU. Therefore, MEC tries to find a good trade-off between Solution A and B with a tunable parameter $T$ in line 8. (In addition, Solution A is available only if $L$ can be used as an auxiliary space, i.e. it is at least as large as $O$). $T$ is a platform-dependent parameter (e.g., on CPU vs. GPU, or on GPU-compute capability), and we found $T$ around 100 to be a good threshold for latest GPUs.

### 3.4. Analysis

In this section, we analyze the memory saving in MEC over im2col-based convolution. The size of the lowered matrix, $L$ in MEC is:

$$i_n o_w i_h k_w k_c \qquad (3)$$

In comparison with the lowered matrix of im2col (see Eq. (2)), there is approximately a factor of $k_h$. For a more exact comparison, let us form their difference $R$.

$$\begin{aligned} R &= i_n k_c (o_h o_w k_h k_w - o_w i_h k_w) \\ &= i_n k_c o_w k_w (o_h k_h - i_h) \\ &= i_n k_c o_w k_w (\frac{i_h - k_h}{s_h} k_h + k_h - i_h) \\ &= i_n k_c o_w k_w (i_h - k_h)(\frac{k_h}{s_h} - 1) \end{aligned} \qquad (4)$$

Since $i_h > k_h$, MEC always reduces memory footprint as long as $k_h > s_h$ (i.e., there is an overlap between kernel instances). Note that in case $k_h \leq s_h$, there is no redundant information to eliminate.

## 4. Experimental Results

We implemented **MEC** for CPU/GPU in C++ with multi-threaded OpenBLAS, OpenMP, and cuBLAS (cuBLAS) using single 32-bit precision. We also implemented a fully parallelized im2col-based convolution on CPU/GPU (Jia, 2014) with the same libraries. We compared **MEC** with other open-source convolution packages in C++, in order to make fair point-by-point comparison and accurately capture the memory-overhead and performance. We downloaded an open-source FFT-based convolution (cuFFT; Theano-FFT) for GPU. We took an open-source Winograd-based convolution (Falcon, 2016) and optimized it to reduce memory-overhead for CPU, and further modified/optimized it for GPU following (Lavin, 2015; Park et al., 2016a). The brief descriptions of the convolution algorithms in this section are as follows:

**Conv.cpu** Conventional im2col-based convolution for CPU with openBLAS/openMP

**Conv.gpu** Conventional im2col-based convolution for GPU with cuBLAS

**Wino.cpu** Winograd-based $F(2\times2, 3\times3)$ convolution for CPU (applicable only when $k_h = k_w = 3$)

**Wino.gpu** Winograd-based $F(2 \times 2, 3 \times 3)$ convolution for GPU (applicable only when $k_h = k_w = 3$)

**FFT.gpu** FFT-based convolution for GPU with cuFFT

**MEC.cpu** MEC for CPU with OpenBLAS/OpenMP

**MEC.gpu** MEC for GPU with cuBLAS

Note that it is performance-critical to combine multiple $sgemm$ calls into a single `cublasSgemmBatched` call in **MEC.gpu**. When modifying/optimizing **Wino.gpu**, we tried to make the best trade-off between parallelism and memory-overhead (i.e., global memory) by utilizing register/shared-memory as much as possible, and ensured experiments representative. Please see Appendix for details on **Wino.gpu** optimization.

For thorough comparison, we built a comprehensive benchmark set consisting of 12 unique convolution layers, cv1-cv12 from various public DNNs (He et al., 2015; Krizhevsky et al., 2012; Sermanet et al., 2013; Simonyan & Zisserman, 2014; Szegedy et al., 2014) as in Table 2. The runtime in our experiments is measured as a wall-clock time by a standard C++ library, running each algorithm 10 times and reporting the average. Our experiments were performed on the two platforms:

**Mobile** Android phone with ARM7 (MSM8960) for user-side inference and training (**mini-bath size=1**)

**Server** Linux server with Intel **CPU** (E5-2680) and Nvidia **GPU** (P100) for inference and training (**mini-bath size=32**)

We present our results in Fig. 4, and made the following summaries:

- (a) plots the factor by which **MEC.cpu** improves memory-overhead and performance over **Conv.cpu**

Table 2. Benchmarks.

| Name | Input $i_h \times i_w \times i_c$ | Kernel $k_h \times k_w \times o_c, s_h(s_w)$ |
|---|---|---|
| cv1 | 227×227×3 | 11×11×96, 4 |
| cv2 | 231×231×3 | 11×11×96, 4 |
| cv3 | 227×227×3 | 7×7×64, 2 |
| cv4 | 224×224×64 | 7×7×64, 2 |
| cv5 | 24×24×96 | 5×5×256, 1 |
| cv6 | 12×12×256 | 3×3×512, 1 |
| cv7 | 224×224×3 | 3×3×64, 1 |
| cv8 | 112×112×64 | 3×3×128, 1 |
| cv9 | 56×56×64 | 3×3×64, 1 |
| cv10 | 28×28×128 | 3×3×128, 1 |
| cv11 | 14×14×256 | 3×3×256, 1 |
| cv12 | 7×7×512 | 3×3×512, 1 |

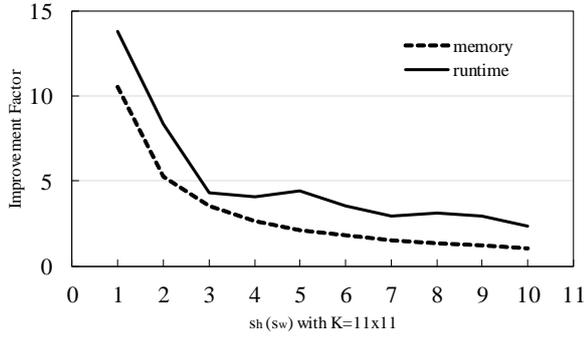

(a) Memory and runtime change for various $s_h = s_w$ values

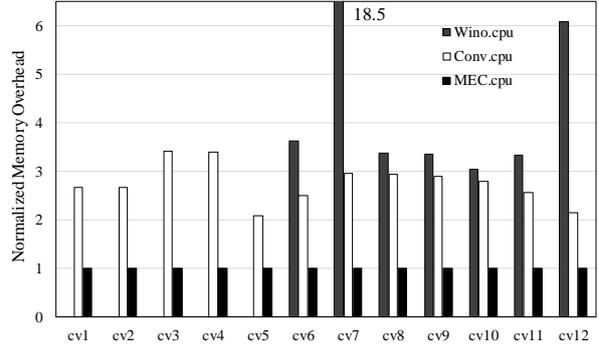

(b) Memory-overhead on **Mobile** and **Server-CPU**

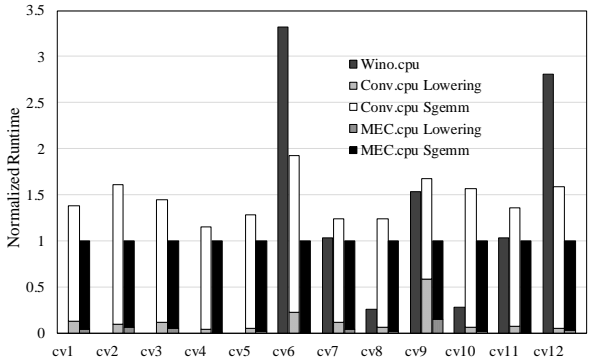

(c) Runtime on **Mobile**

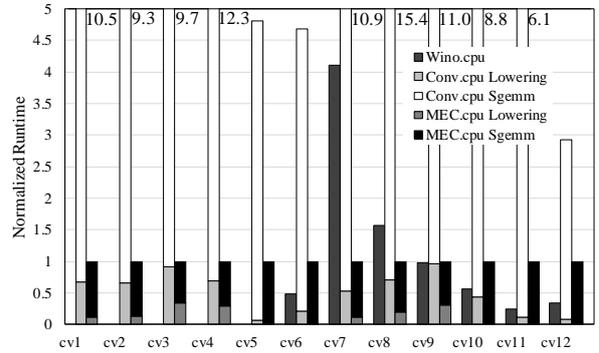

(d) Runtime on **Server-CPU**

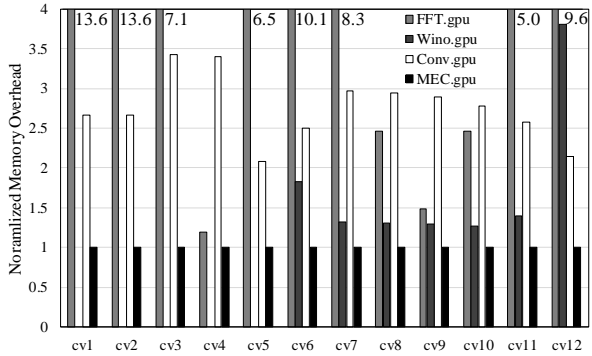

(e) Memory-overhead on **Server-GPU**

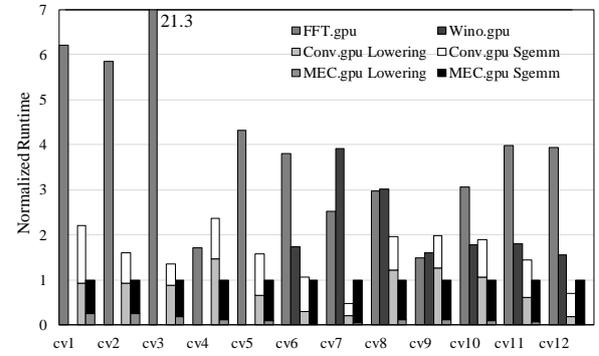

(f) Runtime on **Server-GPU**

*Figure 4.* Memory-overhead and Performance of various sorting convolution algorithms on **Mobile** and **Server**.

for cv1 on **Server-CPU**. While the kernel $K$ is fixed at $11\times11$, $s_h = s_w$ varies from 1 to 10 on the x-axis. We can clearly observe that both memory-overhead and runtime improve with a larger $k/s$ ratio as explained in Eq. (4).

- (b) supports that **MEC** can substantially reduce the memory-overhead. Compared with **Conv.cpu**, the improvement is as large as 3.4x with high $k/s$ ratio, and is on average 3.2x. For cv6-cv12, **MEC.cpu** improves memory-overhead by 5.9x on average, compared with **Wino.cpu**.

- (c) shows that **MEC.cpu** is overall 20% faster than **Conv.cpu** on **Mobile**, yet can be over 90% faster for some layers like cv6. **MEC.cpu** is faster than **Wino.cpu** on 5 benchmarks out of 7.

- (d) shows that on **Server-CPU**, **MEC.cpu** overall shows about 8.8x better runtime than **Conv.cpu**. Compared with **Wino.cpu**, performance is highly de-pendent on the benchmarks: it is similar or faster for cv7,cv8, and cv9.

- (e) presents memory-overheads from various algorithms on **Server-GPU**. **MEC.gpu** shows the least memory-overhead on all benchmarks. **FFT.gpu** requires substantially large memory-overhead. **Wino.gpu** is tested for only cv6-cv12 due to its kernel configuration limitation.

- (f) compares performance of various algorithms on **Server-GPU**. **MEC.gpu** can lower the matrix about 85% faster than **Conv.gpu** due to much fewer bytes to write (which is especially critical on GPU). **Wino.gpu** still has larger memory-overhead than **MEC.gpu** due to the fully parallelized computation of transformed matrices (i.e., $GgG^T$ for each kernel and $B^T dB$ for each channel (Lavin, 2015; Park et al., 2016a)), even though $M$ matrix is kept at registers/shared-memory.

As observed, **MEC** shows greater performance boost on **Server-CPU** than on **Mobile** or **Server-GPU**, because **Server-CPU** is very sensitive to memory-footprint due to the complex cache-architecture. For the example of cv10, we observed through Valgrind cache simulation (Valgrind) that the last-level cache miss in **MEC.cpu** is 0.3%, substantially smaller than 4% in **Conv.cpu**, on a default cache system. **Mobile** has tiny/simple caches, and GPU does not have a sophisticated memory sub-system (deep/big cache hierarchy) to benefit from large memory footprint reduction. Also, cuBLAS is highly optimized to efficiently use fast shared-memory. Overall, **MEC** is all-around player on both **Mobile** or **Server-CPU/GPU** that has no limitation on kernel configuration, incurs the least memory-overhead, yet offers high-performance.

In practice, some convolution layers appear more frequently than others. Therefore, we applied **MEC.cpu** and **Conv.cpu** to ResNet-101 in (He et al., 2015) and estimated the *weighted* impact on memory-overhead and runtime on **Mobile** as in Table 3, which shows that **MEC.cpu** can reduce the memory-overhead by 3x and improve runtime by 20% for a large scale convolutional DNN.

## 5. Conclusion

In this paper, we presented MEC, a memory-efficient convolution algorithm for deep learning. We proposed a novel matrix lowering scheme to improve memory efficiency for MEC which also improves the computational efficiency due to reduced memory footprint. We can clearly observe through extensive experiments that MEC needs the least memory-overhead, yet offers high-performance in most cases on both mobile and server platforms without any restriction, positioning MEC as an attractive convolution engine on various platforms. MEC is well suited for DNN-based applications in memory-constrained environment such as mobile/IoT, while allowing to increase the learning capacity of DNN on high-end server systems.

Table 3. ResNet-101 (He et al., 2015) on **Mobile**.

| | | CONV.CPU | | MEC.CPU | |
|---|---|---|---|---|---|
| NAME | WEIGHT | MEM (MB) | RUNTIME (MSEC) | MEM (MB) | RUNTIME (MSEC) |
| CV4 | 1 | 142.1 | 1228.9 | 41.7 | 1061.3 |
| CV9 | 3 | 19.2 | 26.8 | 6.7 | 16.0 |
| CV10 | 4 | 11.9 | 126.7 | 4.3 | 81.0 |
| CV11 | 23 | 29.1 | 302.7 | 11.3 | 222.9 |
| CV12 | 3 | 1.3 | 16.5 | 0.6 | 10.4 |
| SUM | | 203.6 | 1701.6 | 64.6 | 1391.6 |
| RATIO | | 3.2 | 1.2 | 1.0 | 1.0 |

## Appendix

In this appendix, we sketch **Wino.gpu** optimizations in Section 4 in detail. Our **Wino.gpu** are all hand-tuned/fully-unrolled $F(2 \times 2, 3 \times 3)$ which can fit into the instruction cache in GPU (Lavin, 2015) for maximum performance. We started with an open-source package (Falcon, 2016) and followed the techniques in (Lavin, 2015; Park et al., 2016a) to improve it for GPU. We mainly focused on the high-level optimization including the following:

- For a given input matrix, all transformed kernel and input matrices across all kernels/channels are computed in full parallel for maximum GPU utilization.

- The output matrix is computed by multiplying all pairs of the transformed kernel and input matrices in full parallel for maximum GPU utilization.

- All intermediate products from multiplications are kept in thread registers first and reduced using shared-memory.

- All loops are manually unrolled for maximum performance.

- Read-only cache (`__ldg`) is actively used when computing the output matrix with transformed kernel and input matrices which are shared across blocks.